\documentclass[letterpaper]{article}
\usepackage[preprint]{aaai2027}

\usepackage[hyphens]{url}
\usepackage{graphicx}
\usepackage{natbib}
\usepackage{caption}
\usepackage{amsmath}
\usepackage{amssymb}
\usepackage{amsthm}
\usepackage{algorithm}
\usepackage{algpseudocode}
\usepackage{booktabs}
\usepackage{array}
\usepackage{enumitem}
\usepackage{xcolor}
\usepackage{float}
\usepackage{listings}

\graphicspath{{figures/}{appendix/figures/}}

\lstdefinestyle{tracecadpython}{
    language=Python,
    basicstyle=\ttfamily\footnotesize,
    keywordstyle=\color{blue!55!black},
    commentstyle=\color{green!35!black},
    stringstyle=\color{red!45!black},
    numbers=left,
    numberstyle=\scriptsize\color{black!50},
    numbersep=7pt,
    frame=single,
    rulecolor=\color{black!25},
    breaklines=true,
    showstringspaces=false,
    columns=fullflexible,
    keepspaces=true
}

\title{TraceCAD: Trace-Guided Repair for Agentic CAD Generation}
\author{
    Fengxiao Fan\equalcontrib,
    Jingzhe Ni\equalcontrib,
    Fan Sang,
    Xiaolong Yin,
    Yu Liu,
    Ruofeng Tong,
    Min Tang,
    Peng Du\corresponding
}
\affiliations{
    Zhejiang University, China\\
    dp@zju.edu.cn
}

\newcommand{\method}{TraceCAD}
\newcommand{\stept}{StepTrace}
\newcommand{\cadstep}{CAD step}

\newtheorem{definition}{Definition}

\begin{document}

\maketitle

\begin{abstract}
LLM-based CAD agents produce executable parametric programs, but their correction loops may lose evidence about satisfied requirements, faulty operations, and prior repairs.  We introduce \method{}, a recovery layer that links requested features, modeling steps, failure evidence, and candidate outcomes as persistent state.  \method{} diagnoses likely faulty operations, searches for bounded edits in their dependency regions, validates candidates through execution and preservation checks, and retains successful and failed repair outcomes in reusable skill memory.  On DeepCAD-derived benchmarks with 200-model ablations and a 1K-model comparison, \method{} achieves competitive geometric quality in terms of IoU, Chamfer distance, and Hausdorff distance.  Removing persistent state nearly halves recovery score; removing localized search more than doubles geometric regression and doubles code-agent invocations.  Initializing the skill store on disjoint training models further reduces retries, token cost, and latency.  These results demonstrate that persistent, localized, and reusable recovery improves final CAD quality and repair reliability.
\end{abstract}

\section{Introduction}

Computer-aided design (CAD) generation is shifting from one-shot shape synthesis toward executable modeling.  Recent systems produce editable construction programs from text, images, sketches, or point clouds~\cite{willis2020fusion,wu2021deepcad,xu2022skexgen,text2cad2024,cadllama2025,cadmium2025}.  Code-based methods generate executable CAD scripts that can be rendered, inspected, and iteratively revised~\cite{cadcoder_text2025}.

A nearly correct program may still fail because of an incorrect API call, invalid parameter, unstable operation order, missing feature, or a mismatch between the rendered shape and the request.  Agentic systems use execution logs, visual feedback, and CAD-specific tools to revise such outputs~\cite{cadcodeverify2025,cadassistant2024,seekcad2025,prompt2cad2026,cadialogue2026,cadsmith2026}.  When this evidence remains only in the current prompt, however, later attempts may lack an explicit record of which requirements were satisfied, which operation failed, or which patch should inform future tasks.

CAD programs are structured around modeling operations, and many failures are local even when their visual effects are global.  Recovery should therefore identify a likely responsible step, restrict edits to nearby dependencies, and verify that previously correct features remain unchanged.  Unconstrained regeneration can instead rewrite valid geometry and obscure why a repair worked.

We propose \method{}, a persistent recovery layer for executable CAD agents.  It records feature status together with execution, visual, and repair evidence; diagnoses a likely faulty modeling step and evaluates bounded patches with preservation checks; and converts promoted trajectories into repair skills whose successful and failed reuse outcomes inform later search.  \method{} complements rather than replaces the underlying generator.

\begin{figure*}[t]
    \centering
    \includegraphics[width=0.92\textwidth]{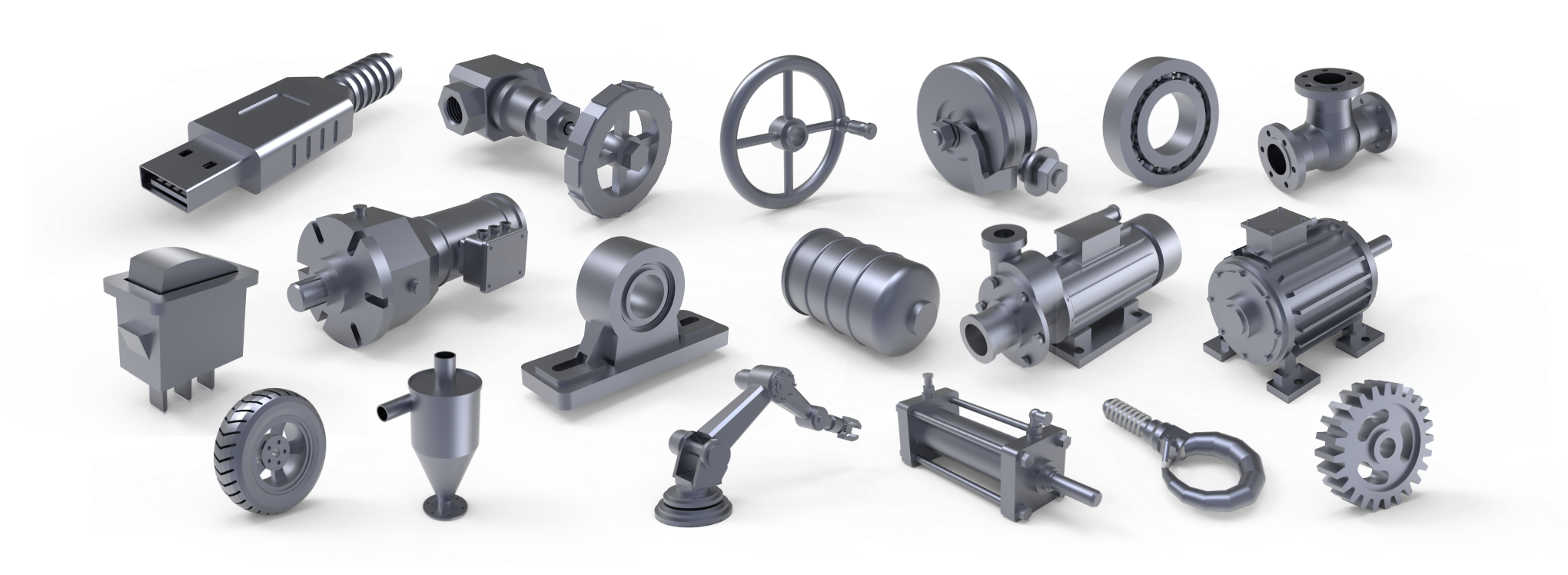}
    \caption{Representative CAD outputs produced by \method{} in independent demonstrations, spanning diverse mechanical parts and multi-component mechanisms.}
    \label{fig:gallery}
\end{figure*}

Figure~\ref{fig:gallery} illustrates the range of executable CAD models produced by \method{}, from individual mechanical parts to more complex multi-component mechanisms.  Our contributions are:
\begin{itemize}
    \item We formulate persistent recovery state for CAD agents, keeping requirements, modeling steps, failures, and repair outcomes inspectable across the correction loop.
    \item We introduce localized repair search for executable CAD programs, combining step-level diagnosis, bounded edit regions, and execute-and-check promotion.
    \item We integrate repair-skill memory into the recovery loop and evaluate both final CAD quality and repair behavior, including recovery score, geometric regression, edit locality, and skill reuse.
\end{itemize}

\section{Related Work}

This section reviews three areas related to our work: parametric CAD generation, tool-augmented CAD agents, and program repair with agent memory.

\subsection{Parametric CAD Generation}

Learning-based CAD generation commonly represents a model as a sequence of construction operations.  Fusion 360 Gallery and DeepCAD established datasets and sequence models for editable histories~\cite{willis2020fusion,wu2021deepcad}.  Later work improves sketch/extrusion structure, hierarchical priors, and boundary-representation generation~\cite{xu2022skexgen,xu2023hierarchical,guo2022complexgen,xu2024brepgen}.  Image-, point-cloud-, and diffusion-based methods further extend structured CAD reconstruction~\cite{ma2024draw,zhang2025diffusion,li2025caddreamer,chen2025cadcrafter}.

Text2CAD maps descriptions to sequential CAD programs, while CAD-Llama and CADmium adapt language models for command generation~\cite{text2cad2024,cadllama2025,cadmium2025}.  CAD-Recode reconstructs code from point clouds, while CAD-Coder generates executable scripts from text~\cite{cadrecode2024,cadcoder_text2025}.  These methods focus primarily on initial generation.  \method{} instead studies how an agent represents and repairs residual failures after proposing a program.

\subsection{Agentic CAD Modeling}

Tool-augmented CAD agents connect LLM reasoning to executable modeling and feedback.  CADCodeVerify and CAD-Assistant use visual verification or CAD-specific tools to iteratively improve executable models~\cite{cadcodeverify2025,cadassistant2024}.  Seek-CAD, CADDesigner, and CADSmith study self-refinement, general-purpose agents, and programmatic geometric validation~\cite{seekcad2025,caddesigner2025,cadsmith2026}.  Prompt2CAD and CADialogue support conversational generation and refinement through textual, visual, or human feedback~\cite{prompt2cad2026,cadialogue2026}.  FlexCAD and CADFusion further explore controllable generation and visual feedback for CAD geometry~\cite{flexcad2025,cadfusion2025}.  Recent systems extend this direction through iterative interaction, solver-grounded operation skills, and expert-derived industrial skills~\cite{itercad2026,embodiedcad2026,artisancad2026}.

These systems establish closed-loop CAD interaction.  \method{} focuses on the complementary problem of representing recovery across attempts and tasks: it ties requirements and failure evidence to local modeling steps, preserved features, candidate outcomes, and reusable repair skills.

\subsection{Program Repair and Agent Memory}

Automated program repair localizes faults, generates patches, and validates them against tests~\cite{genprog2012,tbar2019,xia2023apr}.  LLM agents similarly use reasoning-action traces, tool observations, and self-refinement~\cite{yao2023react,madaan2023selfrefine}.  CAD requires broader validation than compilation or unit tests: a patch must execute, produce a valid artifact, satisfy geometric intent, and preserve unrelated features.

An executable CAD patch can still delete a requested feature, while a global rewrite can discard valid construction history.  \method{} adapts repair to structured design revision by coupling every attempt with step-level evidence, preservation checks, and reuse statistics.  It is designed to make the recovery path local, inspectable, and reusable across tasks.

\section{Method}
\label{sec:method}

\begin{figure*}[t]
    \centering
    \includegraphics[width=0.93\textwidth]{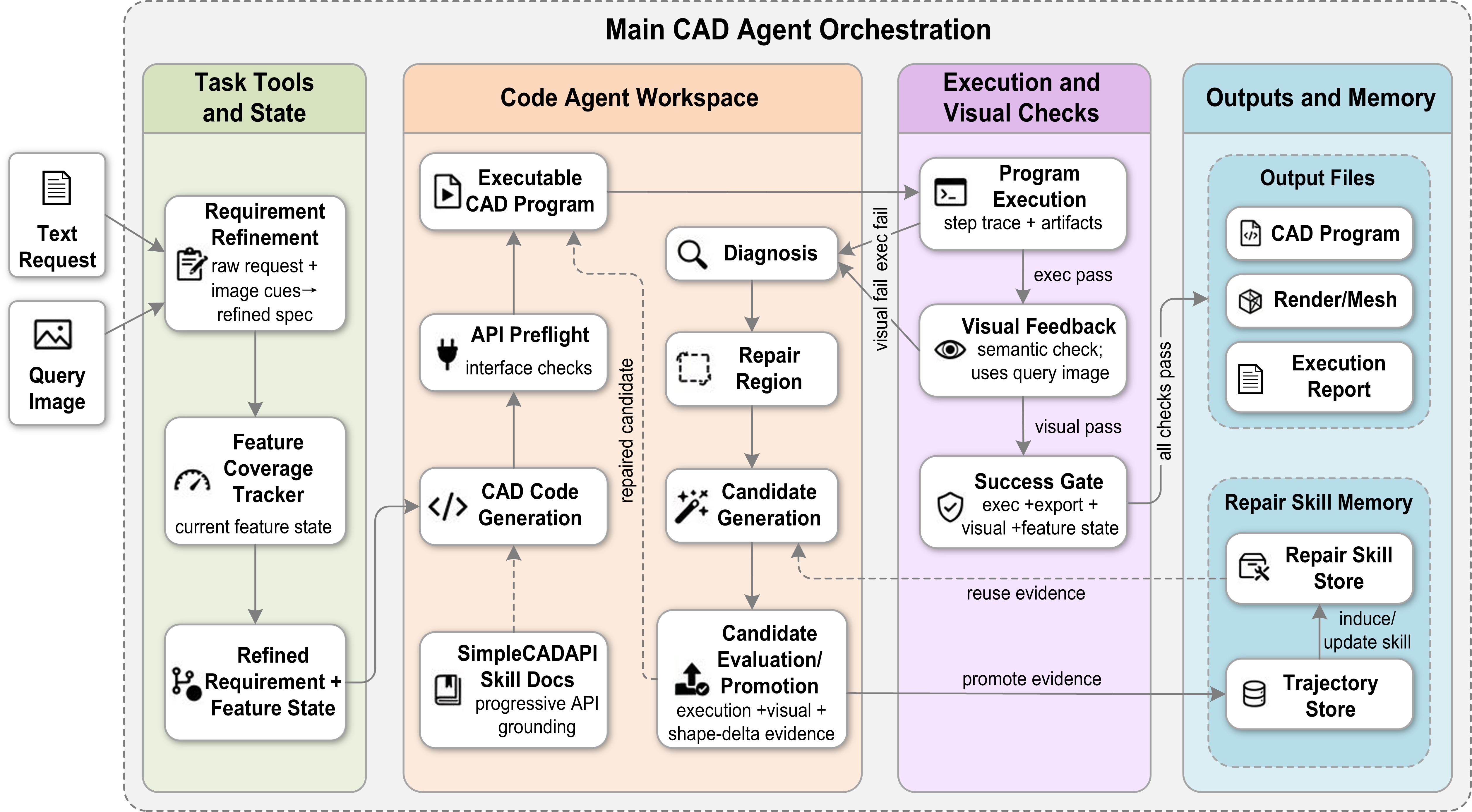}
    \caption{Overview of \method{}. Given a text or multimodal CAD requirement, the agent refines the request, generates executable modeling code, and evaluates the result through execution and visual feedback.  When a failure remains, \method{} records requirement, execution, and visual evidence as persistent recovery state, uses it to diagnose the faulty modeling step, evaluates bounded local repairs, and converts promoted repair trajectories into reusable repair-skill memory.}
    \label{fig:overview}
\end{figure*}

\method{} wraps an executable CAD agent with the recovery loop in Figure~\ref{fig:overview}.  It instruments generated code, records execution and visual evidence, diagnoses a likely faulty modeling step, evaluates local patches, and converts promoted repairs into reusable experience.  The key abstraction is a persistent link between requested features, modeling steps, failure evidence, and candidate outcomes.

Repair is represented as a local, evidence-backed edit rather than another generation attempt.  Persistent records capture what must be preserved, which operation is suspected, which candidates were tested, and why a patch was promoted.  Later decisions can therefore target the suspected operation without reconstructing correct geometry from prompt history.

\subsection{Problem Formulation}

Let \(q\) be a CAD requirement, \(p\) a generated program, and \(F=\{(f_j,z_j)\}\) the persistent feature state, where \(z_j\) records the current lifecycle state of requested feature \(f_j\).  Execution yields report \(e\), artifacts \(a\), visual feedback \(v\), and step trace \(T\).  A run succeeds when the program executes, exports the expected artifacts, and passes visual verification when enabled.  Otherwise, \method{} searches for

\begin{equation}
c^\star =
\arg\max_{c \in \mathcal{B}(\mathcal{C}, d, R, \mathcal{K}, \beta)}
S(c; q,p,F,e,a,v,T),
\end{equation}
where \(\mathcal{C}\) is the set of proposed patches, diagnosis \(d\) identifies the likely faulty step, \(R\) bounds editable code, skill store \(\mathcal{K}\) supplies prior repairs, and \(\beta\) limits search.  The operator \(\mathcal{B}\) restricts \(\mathcal{C}\) to promotable candidates within the diagnosed region and available budget.  \(S\) ranks the remaining candidates by execution, exported artifacts, absence of failed steps, edit locality, and visual score.  Execution is mandatory, while available visual evidence is used to assess the intended change and potential damage to unrelated geometry.  If no candidate passes, no repair is selected; the system retains the current program and retries within budget before returning a failure report.

\subsection{Persistent Recovery State}

The recovery record has two anchors.  Feature state \(F\) represents obligations such as bodies, patterns, edge treatments, and exports; each feature stores its lifecycle state, supporting evidence, related modeling scopes, and open questions.  Step state aligns code with CAD intent by instrumenting feature-level regions:
\begin{lstlisting}
with cad_step("mounting_holes",
    intent="cut four corner mounting holes",
    step_type="hole_pattern"):
    ...
\end{lstlisting}

\begin{definition}[\stept{}]
For a program \(p\) decomposed into scopes \(\{s_i\}_{i=1}^{n}\), a \stept{} is the ordered trace
\[
T=\langle r_1,\ldots,r_n\rangle,
\]
where \(r_i\) records a scope's identity, intent, execution status, timing, and exception evidence.
\end{definition}

Tracebacks thus identify modeling operations rather than only source lines.  Export status, visual feedback, and candidate outcomes update the same record.  An open feature can point to its intended steps, while a failed step explains why the feature remains incomplete or unstable.  This bidirectional link lets later repairs preserve verified features instead of reconstructing state from prompt history.

\subsection{Localized Repair Search}

Execution failures often identify a failed \cadstep{} directly; visual failures are attributed to candidate steps by the vision model using feature names, spatial descriptions, and recorded intents.  The diagnosis records its evidence and a repair family (e.g., API misuse, invalid parameter, missing feature, operation order, topology, export, or visual mismatch).

Because geometry flows across operations, \method{} builds a dependency graph from variables defined and used by each step.  An edge connects a step that produces a value to a later step that uses it.  For suspected step \(s\), the initial repair region is
\begin{equation}
R_h(s)=\{s\}\cup \textsc{Upstream}(s,h),
\end{equation}
where \(h\) is a small hop budget.  Downstream consumers are checked after patching rather than automatically included, which prevents a shared base shape from making every edit global.

The system draws on retrieved repair skills and API documentation disclosed as needed to generate candidates. It uses lightweight checks for interface fixes and the LLM for semantic changes such as adding a pattern or reordering operations, while recording the target scope and rationale for each proposal. To keep edits local, exact text edits and single-step replacements are preferred, with full-program repair retained as a fallback when the change cannot be expressed locally.

Each candidate is then evaluated as a complete program, with successful execution required for promotion and both export status and edit locality used for ranking. When visual checking is enabled, the candidate must additionally pass the visual gate, which uses visual feedback and shape-delta evidence to assess the intended repair and potential damage to unrelated geometry. Candidate promotion therefore remains an intermediate step, while final task completion requires valid exports.

\subsection{Repair Skill Memory}

Each promoted trajectory records the failed program, diagnosis, attempted candidates, accepted patch, and promotion evidence.  An eligible promoted trajectory may then induce a reusable skill:
\begin{equation}
k=(\sigma,d,\pi,\tau,\epsilon,u).
\end{equation}

\begin{figure}[t]
    \centering
    \includegraphics[width=\linewidth]{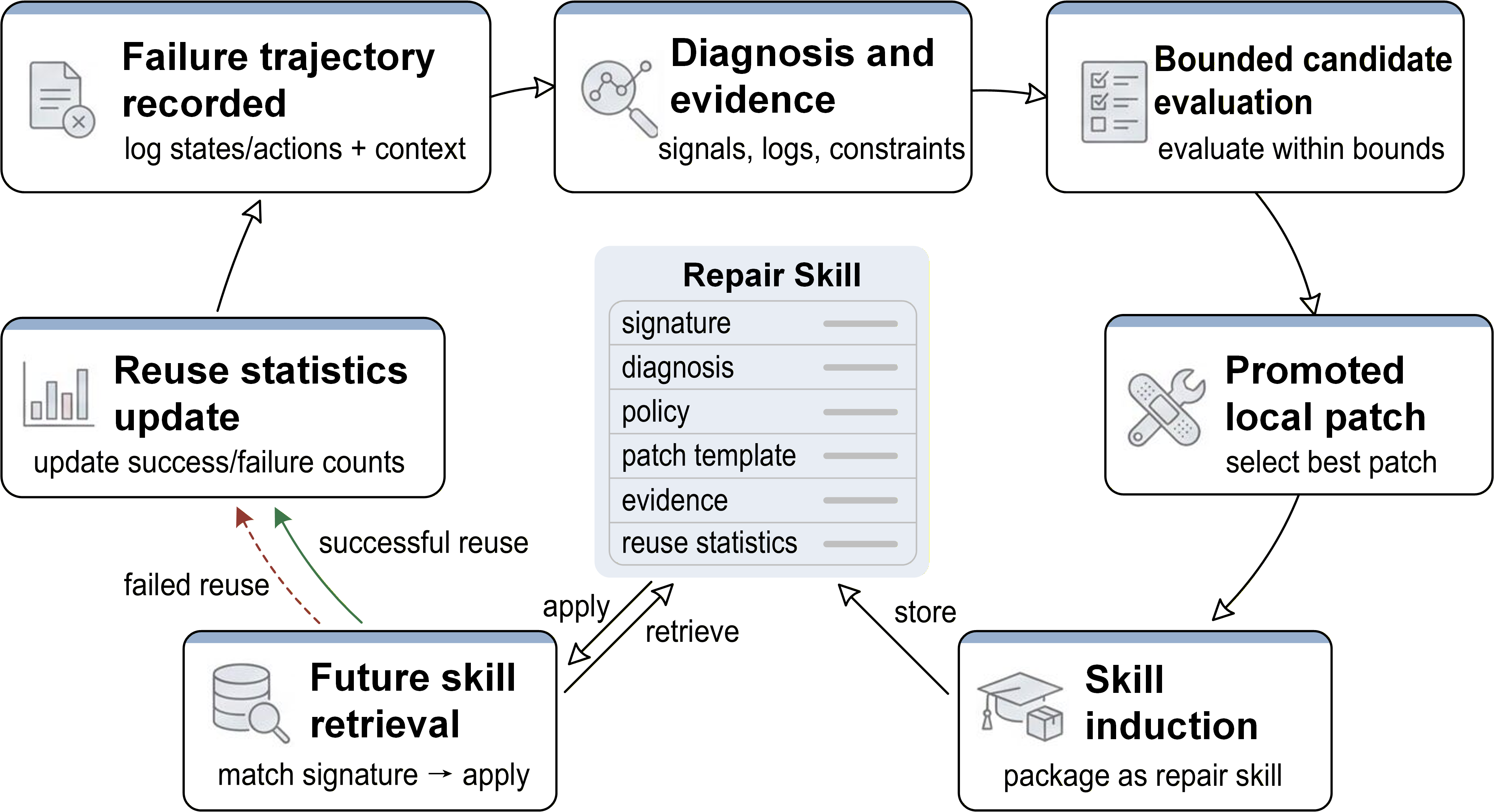}
    \caption{Repair-skill lifecycle.  A failure trajectory is diagnosed through bounded candidate evaluation; an eligible promoted trajectory may induce a repair skill, whose later reuse outcomes update its statistics.}
    \label{fig:skill-lifecycle}
\end{figure}

The fields separate when a skill applies from what it recommends and how its reliability is assessed.  Specifically, \(\sigma\) serves as the retrieval key by describing the circumstances of a failure, including its error terms, implicated step type, repair family, visual symptom, and local context.  \(d\) abstracts the case-specific diagnosis into a transferable failure cause.  Policy \(\pi\) specifies how a candidate should be repaired and which constraints it must respect, while template \(\tau\) instantiates that policy as a minimal reusable patch pattern.  Evidence \(\epsilon\) bundles the execution, artifact, visual, and shape-delta checks that justified promotion.  Finally, \(u\) records retrieval and successful or failed reuse so that later ranking can favor reliable skills.

When a new failure occurs, the system retrieves skills using \(\sigma\) and the current step context, with \(u\) affecting their ranking.  Reuse counts as successful only when a retrieved skill is associated with a promoted repair.  Failed execution, wrong-region edits, preservation violations, and regressions are retained as negative evidence and reduce later ranking.  Figure~\ref{fig:skill-lifecycle} summarizes this lifecycle.

Algorithm~\ref{alg:tracecad} summarizes one bounded repair invocation within the overall recovery loop.  Recovery begins only after initial generation, so the framework can wrap different CAD generators without changing their decoding procedure.  State and skills influence diagnosis and candidate generation, but every reused repair must still pass task-specific validation.

\begin{algorithm*}[t]
\caption{\method{} repair loop after initial code generation}
\label{alg:tracecad}
\begin{algorithmic}[1]
\Require Requirement \(q\), generated program \(p\), skill store \(\mathcal{K}\), budget \(\beta\)
\State Initialize or update feature state \(F\) from \(q\) in main-agent context
\State Run API preflight and execute \(p\) with \cadstep{} instrumentation to obtain \(e,T,a\)
\State Run visual checks when artifacts are renderable to obtain feedback \(v\); update \(F\)
\If{\(e\) succeeds, expected artifacts are exported, and visual verification passes when enabled}
    \State \Return \(p\)
\EndIf
\State Retrieve relevant skills \(\mathcal{K}_r \leftarrow \textsc{SearchSkills}(\mathcal{K},e,v,T)\)
\State Diagnose \(d \leftarrow \textsc{Diagnose}(q,p,e,v,T,\mathcal{K}_r)\)
\State Build repair region \(R \leftarrow \textsc{RepairRegion}(p,d,T)\)
\State Generate and bound candidates \(\mathcal{C}_b \leftarrow \mathcal{B}(\textsc{Generate}(q,p,d,R,\mathcal{K}_r),d,R,\mathcal{K},\beta)\)
\ForAll{\(c \in \mathcal{C}_b\)}
    \State Apply and execute \(c\); score execution, artifact export, locality, and available visual evidence
\EndFor
\State Select the best \(c^\star\) with execution success and, when enabled, visual acceptance
\If{a promotable \(c^\star\) exists}
    \State Update \(F\) and record the trajectory; if eligible, induce or update a repair skill
    \State \Return repaired program
\Else
    \State \Return diagnosis and candidate outcomes
\EndIf
\end{algorithmic}
\end{algorithm*}

\section{System}

\subsection{Agent Runtime}

\method{} is implemented as a two-agent workflow around executable Python CAD programs.  The main agent refines text, image, or mixed input into a task specification and initializes feature coverage.  This state tracks the lifecycle, evidence, and related modeling steps of requested structures.  The code agent receives the specification, retrieves the required SimpleCADAPI documentation~\cite{caddesigner2025}, and generates a program instrumented with \cadstep{} scopes.

Before execution, a lightweight abstract syntax tree (AST) check detects common interface errors, including missing exports, unsupported keyword arguments, and incorrect instrumentation imports.  The executor then runs the program, records its \stept{}, and exports artifacts when possible.  Rendered views of the artifacts are passed to a vision model that estimates their semantic agreement with the request.  The execution report, trace, rendered feedback, and updated feature coverage return to the main agent.

The run terminates after successful execution and export, provided that visual verification also passes when enabled.  Otherwise, the code agent invokes Algorithm~\ref{alg:tracecad}: it retrieves relevant skills, diagnoses a likely faulty step, builds a bounded dependency region, generates patches, and executes each candidate independently.  This separation keeps requirement coverage and visual intent in the main-agent context while API signatures, code edits, traces, and candidate outcomes remain in the code-agent context.

During multi-round interactions, the runtime retains recent history, compresses older tool output, and reintroduces the latest feature state, execution evidence, visual report, and patch outcomes before each decision.  This keeps the active context compact and focused on the current recovery state, limiting token growth and distraction from stale or redundant tool output.

\subsection{Progressive API Grounding}

SimpleCADAPI is a specialized modeling interface, so the agent receives explicit API documentation to ground generation and repair.  CADDesigner retrieves this documentation through RAGFlow and injects the retrieved passages into the agent context~\cite{caddesigner2025}; \method{} instead exposes the same API knowledge through a compact local skill.  Common operations and usage constraints form its entry point, while detailed signatures, examples, and modeling notes are disclosed only when a requested operation needs them.  The same mechanism grounds repair candidates, avoiding a long static API prompt while keeping local edits consistent with the evaluation backend.

\begin{table}[t]
\centering
\small
\begin{tabular*}{\linewidth}{@{\extracolsep{\fill}}p{0.30\linewidth}p{0.60\linewidth}}
\toprule
Record & Runtime role \\
\midrule
Feature coverage & Tracks feature lifecycle states, evidence, related steps, and open questions. \\
\stept{} trace & Stores step intent, status, exception, and timing. \\
Execution/visual reports & Capture program outcome, artifacts, views, and semantic feedback. \\
Candidate reports & Record edit scope, execution result, and rejection cause. \\
Shape delta & Checks intended edits and unintended geometric changes. \\
Repair skills & Store failure signatures, policies, templates, evidence, and reuse statistics. \\
\bottomrule
\end{tabular*}
\caption{Structured evidence maintained during generation and repair.}
\label{tab:runtime-artifacts}
\end{table}

\subsection{Runtime Evidence and Records}

Each attempt produces the structured records summarized in Table~\ref{tab:runtime-artifacts}.  Execution reports identify the failed \cadstep{}, exception, exports, and timing.  Visual reports describe missing features, distorted proportions, or incorrect layout.  Candidate reports record the patch mode, changed scope, execution result, and rejection cause.  Shape-delta inspection asks the vision model to compare rendered artifacts before and after repair and report intended and unintended changes.

The promotion rule consumes these records jointly.  Execution is mandatory, and ranking favors expected exports and edits near the diagnosed region.  When renderings are available, a separate visual gate evaluates candidates using shape-delta judgments or visual-score improvements.  Rejected candidates remain useful because they expose misdiagnosis, unsafe edit scope, or an unreliable retrieved skill.  Together, these records make promotion decisions auditable and preserve the evidence needed to analyze recovery behavior.

\begin{table*}[t]
    \centering
    \small
    \setlength{\tabcolsep}{2.0pt}
    \begin{tabular*}{\textwidth}{@{\extracolsep{\fill}}lcccccccccc}
        \toprule
        & \multicolumn{3}{c}{\textbf{Geometry}} & \multicolumn{4}{c}{\textbf{Repair}} & \multicolumn{3}{c}{\textbf{Task and Efficiency}} \\
        \cmidrule(lr){2-4}\cmidrule(lr){5-8}\cmidrule(lr){9-11}
        \textbf{Variant} &
        \textbf{IoU$\uparrow$} & \textbf{CD$\downarrow$} & \textbf{HD$\downarrow$} &
        \textbf{Rec.$\uparrow$} & \textbf{Reg.$\downarrow$} & \textbf{Scope$\downarrow$} & \textbf{Reuse$\uparrow$} &
        \textbf{AVG Re$\downarrow$} & \textbf{Tokens (K)$\downarrow$} & \textbf{Latency (s)$\downarrow$} \\
        \midrule
        w/o persistent recovery state & 0.3328 & 15.77 & 0.2146 & 0.4872 & 0.5299 & 0.6725 & 20.6\% & 2.3 & 137.9 & 384.6 \\
        w/o localized repair search & 0.3452 & 17.00 & 0.2175 & 0.4865 & 0.7296 & -- & -- & 3.0 & 137.1 & 393.7 \\
        w/o visual feedback & 0.3170 & 18.98 & 0.2204 & 0.7843 & -- & 0.4824 & 9.8\% & 1.4 & 78.2 & 209.2 \\
        w/o skill store & 0.3533 & 15.26 & 0.2193 & 0.4583 & 0.4795 & 0.4776 & -- & 1.9 & 104.6 & 278.9 \\
        \method{} full (cold-start) & \textbf{0.3639} & 10.92 & 0.2002 & \textbf{0.9167} & \textbf{0.3456} & 0.3027 & 31.7\% & 1.5 & 103.6 & 240.8 \\
        \method{} full (warm-up) & 0.3627 & \textbf{10.19} & \textbf{0.1872} & 0.8438 & 0.3460 & \textbf{0.2707} & \textbf{42.9\%} & \textbf{1.2} & \textbf{70.3} & \textbf{177.7} \\
        \bottomrule
    \end{tabular*}
    \caption{Ablation results on the 200-model subset, covering final geometry, repair behavior, and inference efficiency.  A dash indicates that the required evidence is absent for that ablation, such as no promoted local candidate, no skill retrieval, or no post-export repair transition.}
    \label{tab:ablation}
\end{table*}

\section{Experiments}
\label{sec:experiments}

\subsection{Setup}

\method{} is implemented in Python 3.12 and evaluated on a 16-core Intel workstation.  All variants share the SimpleCADAPI backend, executor, renderer, documentation, retry budget, and metric scripts.  The main and code agents use gemini-3.1-pro-preview; visual feedback, candidate scoring, and shape-delta inspection use gemini-3-flash-preview.  The main agent uses temperature 0.1, while code- and vision-model calls use task-specific temperatures from 0.0 to 0.2.  The active context retains the four most recent interaction blocks, each comprising an assistant response and its associated tool results, and compacts older tool output.  Each bounded repair region evaluates at most three candidates spanning at most two target steps; the region starts at one upstream hop and may expand once to two hops.  Skill retrieval returns at most five matches based on signature, step context, and prior reuse outcomes.  When visual evidence is available, promotion requires either a passing shape-delta judgment with no reported unintended changes or a visual-score improvement of at least five points.

We construct benchmarks from DeepCAD~\cite{wu2021deepcad}, removing duplicates and keeping training and test models disjoint as in SkexGen~\cite{xu2022skexgen}.  Following the CADDesigner~\cite{caddesigner2025} protocol, we construct a stratified 200-model subset for ablations and a randomly sampled 1K-model subset for comparison.  Each has a fixed-view rendering, from which we generate a shape-only text description with GPT-5.5 to avoid mismatches in the original text annotations.  Test cases are used only for evaluation and are disjoint from any training models used to initialize repair skills.

\subsection{Metrics}
\label{sec:metrics}

We evaluate three complementary aspects of each method: final geometry, repair behavior, and task-level efficiency.

For geometric fidelity, we report Intersection over Union (IoU), Chamfer Distance (CD), and Hausdorff Distance (HD).  IoU measures volumetric overlap, and HD captures the worst-case point-set discrepancy.  CD values in the tables are multiplied by \(10^3\) for readability.  All meshes and sampled point clouds are normalized to fit within the canonical cube \([-0.5,0.5]^3\).  For pairwise comparison, the generated shape is rigidly aligned to the reference shape using Iterative Closest Point (ICP).  IoU is computed as volumetric BRep overlap after alignment.  CD and HD are computed from 2048 points uniformly sampled from each mesh surface.

For repair evaluation, Recovery Score (Rec.) weights a successful recovery by the inverse number of subsequent code-agent invocations and assigns an unrecovered failure zero.  Regression (Reg.) measures reference-inconsistent geometric change introduced by repair, including lost initially correct geometry and newly added erroneous geometry.  Repair Scope Ratio (Scope) is the normalized AST statement edit distance between the initial program and a promoted repair.  Repair-Skill Reuse Precision (Reuse) is the fraction of retrieved skill uses associated with promoted repairs.

For task-level efficiency, Success Rate (SUC) is the fraction of evaluated cases that execute successfully and export valid STEP and STL artifacts within the retry budget.  Average Retry Count (AVG Re) is the mean number of code-agent invocations before termination, including the initial generation and later repair attempts.  Token Cost (Tokens) measures total model tokens, and End-to-End Latency (Latency) measures elapsed time from task submission to termination.

\subsection{Ablations}
\label{sec:ablation}

We use ablations to separate the effects of recovery state, localized repair search, visual feedback, and repair-skill reuse.  Repair-skill stores are isolated across configurations.  The cold-start setting begins with an empty repair-skill store; eligible promoted trajectories may induce skills online, and those skills may be reused by later tasks in the same evaluation run.  The state ablation removes structured feature and step evidence; the search ablation removes bounded local candidate search and execute-and-promote selection; the skill ablation disables retrieval and induction.  The visual-feedback ablation removes rendered semantic evidence from diagnosis and repair validation, isolating the contribution of visual agreement beyond executable CAD feedback.  A warm-up variant initializes the store on 1K disjoint training models before evaluation.  All variants use text-only input.

Table~\ref{tab:ablation} shows that the components play distinct roles.  Warm-up primarily improves efficiency, while its higher Reuse and lower Scope suggest that transferred experience guides repairs toward smaller, more applicable edits.  Persistent state and localized search are central to reliable recovery: removing either nearly halves Rec., and removing local search additionally produces the most repair attempts and the largest Reg., indicating that bounded candidate evaluation helps avoid broad, damaging revisions.  The skill-store ablation yields the lowest Rec. and a larger repair scope, supporting the value of retaining successful repair experience.  Finally, removing visual feedback reduces cost but produces the weakest geometry, exposing a trade-off between early termination and faithful validation.

\subsubsection{Input Modality}

\begin{table}[t]
    \centering
    \small
    \setlength{\tabcolsep}{3pt}
    \begin{tabular*}{\linewidth}{@{\extracolsep{\fill}}lccccc}
        \toprule
        \textbf{Input} & \textbf{IoU$\uparrow$} & \textbf{CD$\downarrow$} & \textbf{HD$\downarrow$} & \textbf{SUC$\uparrow$} & \textbf{AVG Re$\downarrow$} \\
        \midrule
        Text only & 0.3639 & 10.92 & 0.2002 & \textbf{100.0\%} & \textbf{1.5} \\
        Image only & 0.3865 & 15.38 & 0.2108 & \textbf{100.0\%} & 2.0 \\
        Text + Image & \textbf{0.4020} & \textbf{10.02} & \textbf{0.1838} & \textbf{100.0\%} & 2.2 \\
        \bottomrule
    \end{tabular*}
    \caption{Input-modality ablation on the 200-model subset.}
    \label{tab:modality-ablation}
\end{table}

Table~\ref{tab:modality-ablation} compares \method{} under different input modalities on the 200-model subset.  Image-only input improves IoU over text-only input, suggesting that visual input helps recover global silhouette, proportions, and component layout, but its CD and HD are higher.  A single rendering can leave fine details or hidden topology ambiguous, which may explain these mixed results.  Text and image together give the best geometry, reaching 0.4020 IoU, 10.02 CD, and 0.1838 HD.  This indicates that text identifies intended features and functions, while the image constrains their spatial arrangement.  The multimodal gain comes with additional recovery overhead, increasing AVG Re from 1.5 to 2.2, which suggests that stricter or ambiguous visual feedback can trigger extra repair attempts.

\subsection{Comparison with Existing Methods}
\label{sec:comparison}

We compare \method{} with three representative text-conditioned CAD generation methods on the 1K-model subset using the same evaluation scripts.  Text2CAD~\cite{text2cad2024} is a learning-based model that maps text to sequential CAD programs.  CADCodeVerify~\cite{cadcodeverify2025} is an agentic baseline that iteratively validates and revises generated CAD code using vision-language feedback.  CADDesigner~\cite{caddesigner2025} is a general-purpose ReAct agent that combines requirement refinement, retrieval-grounded code generation, execution, and visual feedback.  We reproduce all baselines under the same benchmark protocol; the two agentic baselines use the same backend model configuration as \method{} for a controlled comparison.

\begin{table}[t]
    \centering
    \small
    \setlength{\tabcolsep}{3pt}
    \begin{tabular*}{\linewidth}{@{\extracolsep{\fill}}lcccc}
        \toprule
        \textbf{Method} & \textbf{IoU$\uparrow$} & \textbf{CD$\downarrow$} & \textbf{HD$\downarrow$} & \textbf{SUC$\uparrow$} \\
        \midrule
        Text2CAD & 0.1141 & 51.46 & 0.3414 & 99.4\% \\
        CADCodeVerify & 0.3136 & 23.74 & 0.2341 & 95.3\% \\
        CADDesigner & 0.3610 & 19.55 & 0.1966 & \textbf{100.0\%} \\
        \method{} full (cold-start) & \textbf{0.3837} & \textbf{14.43} & \textbf{0.1872} & \textbf{100.0\%} \\
        \bottomrule
    \end{tabular*}
    \caption{Comparison on the 1K-model test subset.}
    \label{tab:method-comparison}
\end{table}

\begin{figure}[t]
    \centering
    \includegraphics[width=\linewidth]{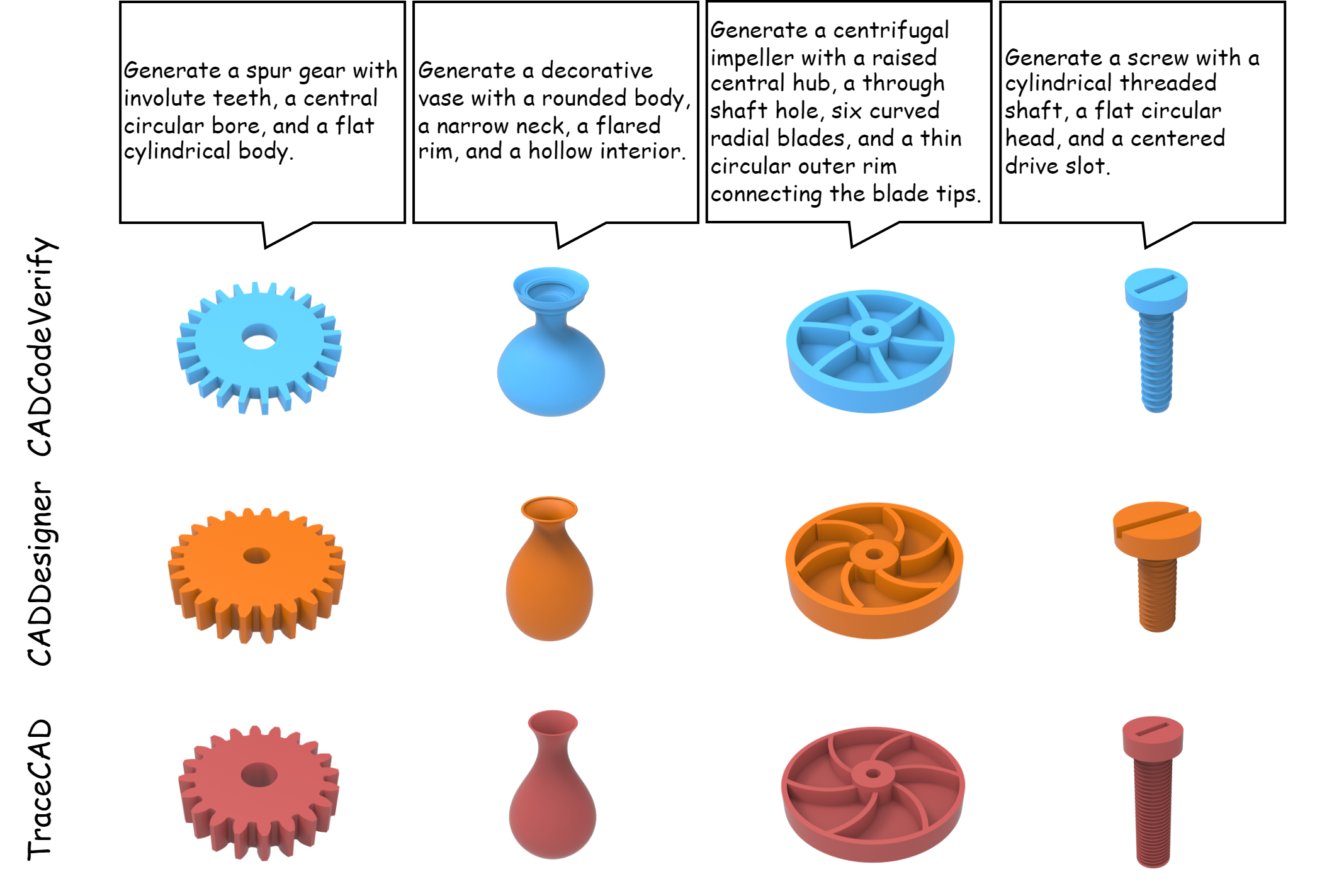}
    \caption{Qualitative results under identical prompts for a gear, vase, impeller, and screw.}
    \label{fig:qualitative-comparison}
\end{figure}

Table~\ref{tab:method-comparison} shows that \method{} full (cold-start) achieves the strongest overall geometric fidelity among the compared methods.  Text2CAD remains highly executable but is less faithful in this abstract text-conditioned setting, while CADCodeVerify improves on Text2CAD but still trails \method{} geometrically.  CADDesigner is the strongest baseline; both CADDesigner and \method{} successfully complete all evaluated cases, while \method{} achieves higher geometric fidelity.  Sharing the evaluation protocol across methods and controlling the backend for the agentic baselines reduces these sources of variation and focuses the comparison on recovery design.

Figure~\ref{fig:qualitative-comparison} provides qualitative comparisons under identical prompts.  All three methods recover the intended part categories, but the baselines simplify some local construction details: CADCodeVerify produces block-like gear teeth and nearly straight impeller ribs, while CADDesigner captures more of the curved geometry but shows less consistent feature proportions.  In these examples, \method{} preserves the prompt-specified local features, including the central gear bore, hollow flared vase opening, curved impeller blades connected by a continuous outer rim, and threaded shaft with a centered slot, while maintaining the global part structure.  This pattern reflects \method{}'s stronger ability to preserve local CAD features and is consistent with its higher overall geometric fidelity.
\section{Discussion and Limitations}

\method{} targets residual failures after initial generation, using persistent state, localized search, and skill memory to address API errors, kernel failures, missing features, ordering problems, and edit regressions.

This framing exposes behavior hidden by final-shape metrics.  Recovery records distinguish fixing the intended defect from finding another executable program.  The distinction matters in CAD because syntactically valid outputs can remain semantically incomplete or lose editable construction history.

Several limitations remain.  Diagnosis depends on meaningful program structure, so poorly decomposed code makes step attribution ambiguous.  Rendered feedback and shape-delta judgments are model-based semantic evidence rather than formal geometric proof.  Sparse promoted trajectories can make skills overly specific.  Moreover, online cold-start accumulation introduces cross-case dependence and sensitivity to task order and parallel scheduling.

Future work should recover step boundaries automatically, combine local and global repair policies, strengthen geometric validators, and learn skill retrieval from broader repair histories.

\section{Conclusion}

We introduced \method{}, a persistent recovery framework for agentic CAD generation.  It records correction as inspectable state, uses that state for localized repair search, and converts promoted repairs into reusable skills.  The evaluation measures both final CAD quality and the recovery process, including recovery quality, locality, geometric regression, and reuse.

\method{} complements initial CAD generation with persistent recovery.  Rejected candidates and failed skill reuse remain as negative evidence that can down-rank unreliable skills in later searches.  The results suggest that persistent recovery state can complement advances in CAD generation and validation by addressing residual failures.

\bibliography{references}

\clearpage
\onecolumn
\appendix
\setcounter{secnumdepth}{2}

\begin{center}
    {\LARGE\bfseries Supplementary Material}
\end{center}

This supplementary material presents formal metric definitions, additional
experiments on LLM backends and inference cost, and an end-to-end case study
of generation and repair.

\section{Evaluation Metric Definitions}
\label{supp:sec:metrics}

We provide formal definitions for all metrics reported in the main paper. Let
\(N\) denote the number of evaluated cases. For case \(i\), \(G_i\) and \(S_i\)
denote the generated and reference BRep solids, and \(X_i\) and \(Y_i\) denote
point clouds uniformly sampled from their surfaces. Before comparison, each
shape is centered and isotropically scaled to fit within \([-0.5,0.5]^3\). The
generated shape is then rigidly aligned to the reference using
Iterative Closest Point (ICP). After alignment, the generated shape is centered
and isotropically rescaled once more to fit the same canonical cube before
metric evaluation. Unless stated otherwise, each dataset-level result is the
arithmetic mean of the corresponding per-case metric over cases with the
required evidence. Throughout this section, \(|\cdot|\) denotes the cardinality
of a finite set.

For ICP, we sample at most 2048 surface points from each shape and use
point-to-point estimation with identity initialization and a maximum
correspondence distance of \(0.2\) in normalized coordinates. We use the
default Open3D convergence criteria. If alignment fails, the normalized
prediction is retained with the identity transform rather than excluding the
pair.

\subsection{Geometric Fidelity}

\paragraph{BRep Intersection over Union (IoU).}
We compute IoU from the volumes of the aligned BRep solids, where
\(\operatorname{Vol}(\cdot)\) denotes enclosed volume:
\begin{equation}
    \operatorname{IoU}_i
    = \frac{\operatorname{Vol}(G_i \cap S_i)}
    {\operatorname{Vol}(G_i \cup S_i)}
    = \frac{\operatorname{Vol}(G_i \cap S_i)}
    {\operatorname{Vol}(G_i)+\operatorname{Vol}(S_i)
      -\operatorname{Vol}(G_i \cap S_i)}.
    \label{supp:eq:iou}
\end{equation}
The CAD kernel computes the BRep intersection and evaluates its enclosed
volume. Let \(\mathcal{V}_{\mathrm{IoU}}\) denote the set of valid BRep pairs
and \(N_{\mathrm{IoU}}=|\mathcal{V}_{\mathrm{IoU}}|\). We report
\(N_{\mathrm{IoU}}^{-1}\sum_{i\in\mathcal{V}_{\mathrm{IoU}}}
\operatorname{IoU}_i\).

\paragraph{Chamfer Distance (CD).}
For surface point sets \(X_i\) and \(Y_i\), the symmetric squared Chamfer
distance is
\begin{equation}
    \operatorname{CD}_i(X_i,Y_i)
    = \frac{1}{|X_i|}\sum_{x\in X_i}
        \min_{y\in Y_i}\lVert x-y\rVert_2^2
       +\frac{1}{|Y_i|}\sum_{y\in Y_i}
        \min_{x\in X_i}\lVert y-x\rVert_2^2.
\label{supp:eq:cd}
\end{equation}
We sample 2048 points uniformly from each surface. Reported CD values are
multiplied by \(10^3\) for readability.

\paragraph{Hausdorff Distance (HD).}
HD captures the largest bidirectional nearest-neighbor discrepancy:
\begin{equation}
    \operatorname{HD}_i(X_i,Y_i)=\max\bigg\{
    \max_{x\in X_i}\min_{y\in Y_i}\lVert x-y\rVert_2,
    \max_{y\in Y_i}\min_{x\in X_i}\lVert y-x\rVert_2
    \bigg\}.
\label{supp:eq:hd}
\end{equation}
CD measures average surface agreement, whereas HD exposes the most severe local
discrepancy.

\subsection{Repair Behavior}

Let \(\mathcal{F}\) denote the cases whose initial generation fails. For
\(i\in\mathcal{F}\), let \(a_i\geq 1\) be the number of subsequent code-agent
invocations, and let \(z_i\in\{0,1\}\) indicate whether the case ultimately
exports valid STEP and STL artifacts.

\paragraph{Recovery Score (Rec.).}
Rec. discounts successful recovery by the number of repair attempts and assigns
zero to an unrecovered case:
\begin{equation}
    \operatorname{Rec}
    =\frac{1}{|\mathcal{F}|}
      \sum_{i\in\mathcal{F}}\frac{z_i}{a_i}.
    \label{supp:eq:recovery}
\end{equation}
A recovery completed by the first subsequent code-agent invocation therefore
contributes \(1\), one completed by the second contributes \(1/2\), and an
unrecovered failure contributes \(0\).

\paragraph{Geometric Regression (Reg.).}
Reg. measures whether repair removes reference-consistent geometry or introduces
new reference-inconsistent geometry. Let \(B_i\), \(A_i\), and \(R_i\) denote
the occupied voxel sets of the pre-repair baseline, repaired output, and
reference, respectively. We use the first successfully exported STEP model as
the baseline when available; otherwise, we use the executable initial model.
The normalization and reference alignment estimated for the repaired model
are applied to both generated states. Alignment sampling uses a deterministic
seed derived from the case identifier. With the default regression resolution
of 64, occupied voxels are computed with pitch \(2/64\) in normalized
coordinates. Correct Geometry Retention (CGR) and New False Geometry (NFG) are
\begin{equation}
    \operatorname{CGR}_i
    =\frac{|A_i\cap B_i\cap R_i|}{|B_i\cap R_i|},
    \qquad
    \operatorname{NFG}_i
    =\frac{|A_i\setminus(B_i\cup R_i)|}{|R_i|}.
    \label{supp:eq:cgr-nfg}
\end{equation}
The per-case regression score and its dataset mean are
\begin{equation}
    \operatorname{Reg}_i=(1-\operatorname{CGR}_i)
    +\operatorname{NFG}_i,
    \qquad
    \operatorname{Reg}=\frac{1}{|\mathcal{V}_{\mathrm{Reg}}|}
    \sum_{i\in\mathcal{V}_{\mathrm{Reg}}}\operatorname{Reg}_i,
    \label{supp:eq:regression}
\end{equation}
where \(\mathcal{V}_{\mathrm{Reg}}\) contains cases with distinct executable
pre- and post-repair artifacts and nonempty reference-consistent baseline
geometry. Lower values indicate less geometric regression.

\paragraph{Repair Scope Ratio (Scope).}
For promoted candidate \(j\), let \(E_j\) be the unit-cost Levenshtein distance
between the initial and candidate programs' AST statement sequences, with
lengths \(N_0\) and \(N_j\). Formatting, docstrings, and \cadstep{}
instrumentation are excluded. We define
\begin{equation}
    \operatorname{Scope}_j=\frac{E_j}{\max(N_0,N_j)},
    \qquad
    \operatorname{Scope}=\frac{1}{|\mathcal{P}|}
    \sum_{j\in\mathcal{P}}\operatorname{Scope}_j,
    \label{supp:eq:scope}
\end{equation}
where \(\mathcal{P}\) denotes promoted repairs with valid code snapshots
and a nonzero denominator. Lower values indicate more localized edits.

\paragraph{Repair-Skill Reuse Precision (Reuse).}
Let \(n_{\mathrm{ret}}\) denote the number of unique uses of retrieved skills
across localized repair rounds, and let \(n_{\mathrm{prom}}\) denote the number
of those uses associated with promoted repairs. Reuse precision is
\begin{equation}
    \operatorname{Reuse}=\frac{n_{\mathrm{prom}}}{n_{\mathrm{ret}}}.
    \label{supp:eq:reuse}
\end{equation}
The metric measures whether retrieved skill uses are associated with accepted repairs,
rather than whether retrieval merely occurs.

\subsection{Task Success and Efficiency}

Let \(q_i\in\{0,1\}\) indicate whether case \(i\) terminates with valid STEP and
STL artifacts. Let \(c_i\) be its number of code-agent invocations, including
the initial generation and subsequent repairs.

\paragraph{Success Rate (SUC).}
SUC is the fraction of evaluated cases that produce valid STEP and STL
artifacts:
\begin{equation}
    \operatorname{SUC}=\frac{1}{N}\sum_{i=1}^{N}q_i.
    \label{supp:eq:suc}
\end{equation}

\paragraph{Average Retry Count (AVG Re).}
AVG Re counts the initial generation as one invocation:
\begin{equation}
    \operatorname{AVG\ Re}=\frac{1}{N}
    \sum_{i=1}^{N}c_i.
    \label{supp:eq:avg-re}
\end{equation}

\paragraph{Tokens.}
Let \(u_{i\ell}\) denote the total token usage recorded for model call \(\ell\)
in case \(i\), and let \(\mathcal{L}_i\) denote the set of recorded model calls
for that case. The average token cost is
\begin{equation}
    \operatorname{Tokens}=\frac{1}{N}
    \sum_{i=1}^{N}\sum_{\ell\in\mathcal{L}_i}u_{i\ell}.
    \label{supp:eq:tokens}
\end{equation}
The total includes all recorded main-agent,
code-agent, and vision-model calls, including cached-input and reasoning-token
usage when available in the call metadata. Tables labeled ``Tokens (K)'' divide
the result by \(10^3\).

\paragraph{End-to-End Latency.}
Let \(t_i^{\mathrm{start}}\) and \(t_i^{\mathrm{end}}\) denote task submission
and termination times, respectively. Then
\begin{equation}
    \operatorname{Latency}=\frac{1}{N}
    \sum_{i=1}^{N}
    \left(t_i^{\mathrm{end}}-t_i^{\mathrm{start}}\right).
    \label{supp:eq:latency}
\end{equation}
Latency covers model calls, code execution, rendering,
validation, and repair.

\paragraph{Metric availability.}
We do not impute metrics when their required evidence is absent. Scope requires
a promoted candidate and paired code snapshots, Reuse requires a retrieved
skill, and Reg. requires comparable pre- and post-repair geometry. A dash
therefore denotes an unavailable measurement rather than zero.

\section{Additional Experiments}
\label{supp:sec:additional-experiments}

\subsection{Benchmark Construction}
\label{supp:sec:benchmark-construction}

We construct both evaluation subsets from the held-out DeepCAD test split.
Deduplication follows the SkexGen preprocessing pipeline: each construction
history is normalized, its sketch and extrusion sequences are quantized with
six-bit precision, and the concatenated sequence is hashed with SHA-256.
Histories with the same sketch--extrusion hash form a duplicate group, from
which only the lexicographically smallest model identifier is retained.
Models without a valid converted construction history are also excluded from
the sampling pool.

We measure model complexity by the number of sketch primitives plus extrusion
operations. The fixed stratified 200-model subset contains 40 successfully
converted models from each of five command-count intervals: 1--10, 11--20,
21--30, 31--40, and 41 or more. Within each stratum, cases are selected from a
pseudorandom candidate ordering generated with seed 20260705. The 1K subset
is obtained by shuffling the same eligible test pool with the same seed but
without command-count balancing. When a selected history cannot be converted
and exported as a nonempty STEP model, the next candidate in the seeded order
is used.

Each reference STEP model is rendered once using our fixed-view rendering
script. The script samples up to 500 points from the solid interior and applies
PCA to estimate three object-aligned axes. It then forms an oblique viewing
direction from the first two principal axes at a fixed $30^{\circ}$ azimuth
and adds the third axis with weight $\sqrt{2}/2$. This construction exposes
multiple principal extents while avoiding alignment with any single principal
axis. The camera targets the bounding-box center and fits the complete model
into a $496\!\times\!473$ viewport with an opaque white background. We use
GPT-5.5 to produce the shape-only text description from this single image.
This dataset rendering is distinct from the six-view runtime validation
described below.

\subsection{Implementation Settings}
\label{supp:sec:implementation-settings}

All experiments use Python 3.12 and SimpleCADAPI 2.0.0b3. Unless an ablation
explicitly changes a component, the implementation settings in
Table~\ref{supp:tab:implementation-settings} are held fixed. We use
\emph{retry budget} as an umbrella term for these agent-level termination and
repair limits.
Dataset sampling and replacement order use the fixed seed 20260705. The hosted
LLM endpoints do not provide a consistently enforceable sampling seed, so
generation remains stochastic under the fixed model and temperature settings.
Each configuration is executed once per selected case, and the reported
aggregates are computed across cases.

\begin{table}[H]
    \centering
    \small
    \setlength{\tabcolsep}{5pt}
    \begin{tabular*}{\textwidth}{@{\extracolsep{\fill}}
        p{0.25\textwidth}p{0.68\textwidth}@{}}
        \toprule
        \textbf{Setting} & \textbf{Value} \\
        \midrule
        CAD environment & Python 3.12; SimpleCADAPI 2.0.0b3. \\
        Agent step limits & At most 30 main-agent tool steps and 20 internal
        Code Agent steps per invocation. \\
        Repair limits & Bounded search evaluates at most three candidates
        over at most two target
        steps; its dependency region starts at one upstream hop and may expand
        once to two hops. \\
        Execution and stopping & Each program execution has a 180-second
        timeout. The main loop terminates after five consecutive unsuccessful
        repair evaluations or five consecutive visual-feedback failures. \\
        Validation rendering & Six orthographic views at $512\!\times\!512$ pixels,
        arranged in a $3\!\times\!2$ grid with bounds-adaptive framing. \\
        Context management & Four recent interaction blocks, each comprising an
        assistant response and its associated tool results, are retained
        verbatim; older blocks are compacted. Tool context and execution-trace
        previews are capped at 4,000 and 1,200 characters, respectively. \\
        \bottomrule
    \end{tabular*}
    \caption{Common implementation settings and retry-budget limits.}
    \label{supp:tab:implementation-settings}
\end{table}

\subsection{Effect of the LLM Backend}
\label{supp:sec:llm-backend}

We evaluate the sensitivity of \method{} to the underlying LLM by substituting
the backend used for the main-agent, code-agent, and vision-model roles while
keeping the benchmark, SimpleCADAPI environment, retry budget, and metric
implementation fixed. Each backend is evaluated on the same 200 cases used for
the main ablations.
The three backends are comparably capable general-purpose LLMs from different
model families. Gemini-3.1-Pro-Preview is used for the main-agent and code-agent
roles in the main-paper results, with Gemini-3-Flash-Preview for vision.
GPT-5.4 and Claude-Sonnet-4.6 serve all three roles to test whether the findings
transfer to other model families at a similar capability level.

\begin{table}[H]
    \centering
    \small
    \setlength{\tabcolsep}{3pt}
    \begin{tabular*}{\textwidth}{@{\extracolsep{\fill}}lccccccc@{}}
        \toprule
        \textbf{LLM Backend} & \textbf{IoU$\uparrow$} &
        \textbf{CD$\downarrow$} & \textbf{HD$\downarrow$} &
        \textbf{SUC$\uparrow$} & \textbf{AVG Re$\downarrow$} &
        \textbf{Tokens (K)$\downarrow$} & \textbf{Latency (s)$\downarrow$} \\
        \midrule
        GPT-5.4 & 0.3605 & 11.30 & 0.1882 & \textbf{100.0\%} & \textbf{1.3} & \textbf{82.0} & \textbf{223.6} \\
        Claude-Sonnet-4.6 & 0.3596 & 12.09 & \textbf{0.1826} & \textbf{100.0\%} & 1.7 & 100.6 & 297.7 \\
        Gemini-3.1-Pro-Preview & \textbf{0.3639} & \textbf{10.92} & 0.2002 & \textbf{100.0\%} & 1.5 & 103.6 & 240.8 \\
        \bottomrule
    \end{tabular*}
    \caption{Effect of the LLM backend on the 200-model subset.}
    \label{supp:tab:llm-backend}
\end{table}

Table~\ref{supp:tab:llm-backend} shows that the LLM backend affects both geometry
and inference efficiency, but the overall variation in geometry among the
evaluated models remains limited. All three backends reach 100.0\% success, and
their IoU values span only 0.0043. Gemini-3.1-Pro-Preview achieves the best IoU
(0.3639) and CD (10.92), whereas Claude-Sonnet-4.6 obtains the lowest HD
(0.1826). GPT-5.4 requires the fewest code-agent invocations and has the lowest
token cost and latency, while all three backends preserve similar geometric
fidelity. Backend choice therefore has a measurable effect, particularly on
cost, but the limited variation suggests that \method{} does not depend
strongly on the choice among these three backends in this experiment.

\subsection{Inference-Cost Comparison}
\label{supp:sec:cost}

We compare four agent configurations on the same 200 cases, using GPT-5.4 for
all model calls. They share the CAD backend, hardware, and
artifact checks, and differ in agent organization and knowledge access.

\begin{itemize}[leftmargin=*]
    \item \textbf{Single agent} performs planning, API selection, code
    generation, execution diagnosis, and repair within one agent history. It
    retains the local API skill but removes the code-agent boundary, isolating
    the effect of role separation.
    \item \textbf{RAGFlow knowledge injection} replaces the compact local API
    skill with documents retrieved from RAGFlow. This variant evaluates the
    cost of repeatedly retrieving and injecting API documentation instead of
    using the compact local skill.
    \item \textbf{\method{} full} uses main/code-agent separation, a local API
    skill, persistent recovery state, and bounded localized repair.
    \item \textbf{CADDesigner} uses its RAGFlow-based knowledge retrieval and is
    reproduced with the same GPT-5.4 backend and evaluation environment as an
    external agentic baseline.
\end{itemize}

For the RAGFlow-based configurations, retrieval first considers 30 candidate
chunks and returns at most three results per query. We use a page size of 33, a
similarity threshold of 0.2, and a vector-similarity weight of 0.6. No
additional document-count cap is applied after mapping retrieved chunks back
to their source documents.

\begin{figure}[H]
    \centering
    \includegraphics[width=0.85\linewidth]{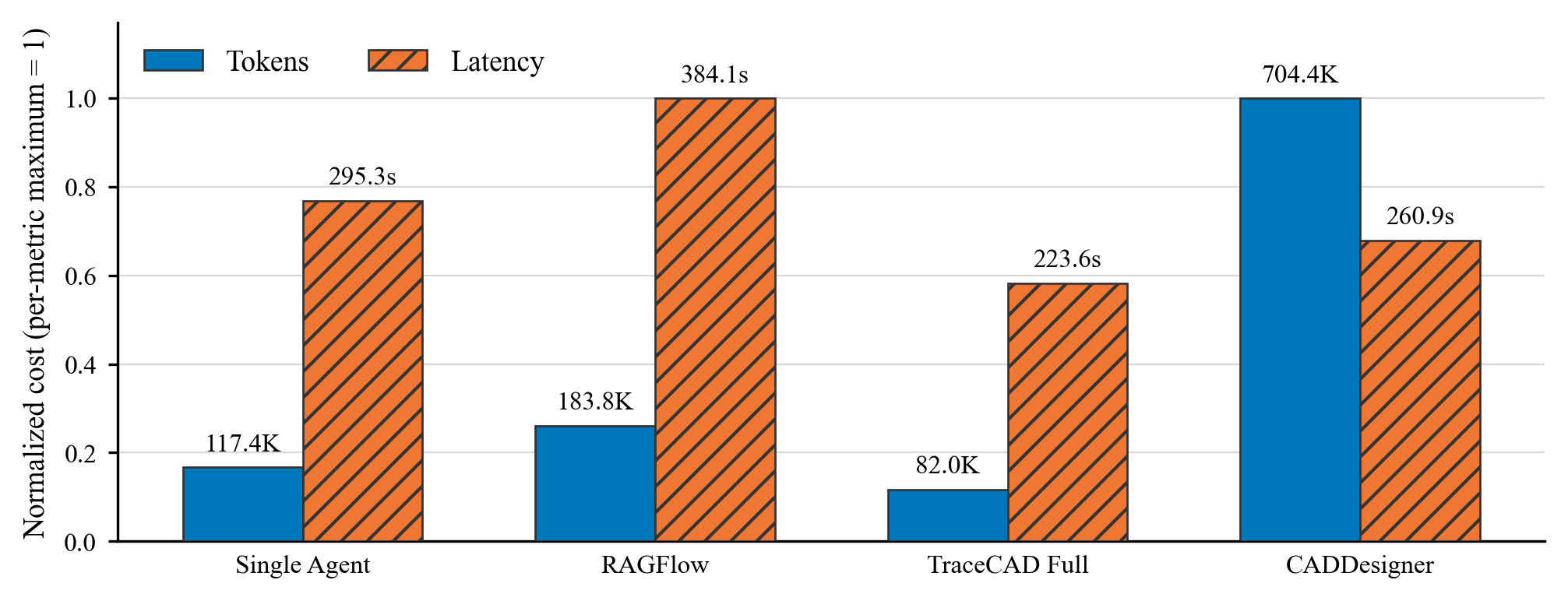}
    \caption{Average token usage and end-to-end latency on the 200-model subset.
    Each metric is normalized independently by its maximum; bar labels report
    the original values in thousands of tokens and seconds. All configurations
    use GPT-5.4.}
    \label{supp:fig:cost-comparison}
\end{figure}

For each configuration, token usage is the mean total token count per case,
and latency is the mean end-to-end wall-clock time in seconds. Both statistics
are computed over all 200 cases rather than only successful cases.
Figure~\ref{supp:fig:cost-comparison} compares the resulting token usage and
latency.

The single-agent configuration retains planning, documentation, execution, and
repair evidence throughout one long-lived context. RAGFlow supplies API
documentation through retrieval but may inject larger passages at several
generation steps, while CADDesigner relies on repeated interaction between its
primary agent and tools. In contrast, \method{} delegates code-local work to a
bounded code agent, retrieves only the relevant API skill content, preserves a
compact recovery state, and executes candidates within a local repair region.
\method{} full reduces mean token usage by 30.1\% relative to the single-agent
variant and by 55.4\% relative to RAGFlow knowledge injection; the corresponding
latency reductions are 24.3\% and 41.8\%. Relative to CADDesigner, the token
reduction is much larger (88.4\%) than the latency reduction (14.3\%). This
smaller latency margin is expected because \method{} performs persistent
recovery, repeated artifact execution, rendering, and stricter visual
validation after code generation. These operations add wall-clock time without
proportionally expanding the language-model context. Thus, part of the time
saved by compact local skills is spent on validation and repair.

\subsection{Paired Statistical Analysis}
\label{supp:sec:paired-tests}

We complement the reported geometry means with two-sided paired Wilcoxon signed-rank
tests on the 1K subset, matching cases by identifier and retaining finite paired
measurements for each metric. Paired differences are
oriented so that positive values favor \method{}: for IoU they are computed as
\method{} minus the baseline, whereas for distance metrics they are
computed as the baseline minus \method{}. We report Holm-adjusted \(p\)-values
together with matched-pairs rank-biserial effect sizes \(r_{\mathrm{rb}}\).
Holm correction is applied across all nine geometry comparisons.

\begin{table}[H]
    \centering
    \small
    \setlength{\tabcolsep}{7pt}
    \begin{tabular*}{\textwidth}{@{\extracolsep{\fill}}lccc@{}}
        \toprule
        \textbf{Baseline} & \textbf{IoU} &
        \textbf{CD} & \textbf{HD} \\
        \midrule
        CADDesigner &
        \(\mathbf{0.001724}\;(0.115)\) &
        \(\mathbf{7.42{\times}10^{-29}}\;(0.411)\) &
        \(\mathbf{0.001263}\;(0.125)\) \\
        CADCodeVerify &
        \(\mathbf{3.30{\times}10^{-32}}\;(0.452)\) &
        \(\mathbf{1.84{\times}10^{-60}}\;(0.617)\) &
        \(\mathbf{2.15{\times}10^{-37}}\;(0.483)\) \\
        Text2CAD &
        \(\mathbf{6.92{\times}10^{-119}}\;(0.875)\) &
        \(\mathbf{2.53{\times}10^{-124}}\;(0.882)\) &
        \(\mathbf{3.64{\times}10^{-112}}\;(0.837)\) \\
        \bottomrule
    \end{tabular*}
    \caption{Paired geometry comparisons. Each entry reports the
    Holm-adjusted \(p\)-value followed by \(r_{\mathrm{rb}}\) in parentheses;
    positive effect sizes favor \method{}.}
    \label{supp:tab:paired-geometry}
\end{table}

Table~\ref{supp:tab:paired-geometry} shows significant improvements over all three
baselines on IoU, CD, and HD after Holm correction across all nine comparisons.
The paired effect sizes are largest against Text2CAD across all three metrics,
whereas the comparisons with CADCodeVerify and CADDesigner show stronger effects
for CD than for IoU or HD.

\section{End-to-End Generation and Repair}
\label{supp:sec:case-study}

We illustrate the complete workflow using a geared electric motor. The
reference image depicts a stepped cylindrical housing, top terminal box,
faceted reduction gearbox, slotted output flange, splined shaft, cable-entry
bosses, and service plugs. The user provides this image with the concise
instruction:
\begin{quote}
\emph{Create a geared electric motor with a cylindrical motor body, a top
terminal box, a polygonal reduction gearbox, a slotted circular output flange,
and a projecting stepped shaft.}
\end{quote}
This case traces requirement refinement, feature tracking, API grounding,
localized repair, and final visual validation.

\subsection{Refined Requirements}

Image-conditioned refinement converts the short request into the operational
specification in Table~\ref{supp:tab:refined-requirements}. The dimensions are
those used by the final validated program.

\begin{table}[H]
    \centering
    \small
    \renewcommand{\arraystretch}{1.15}
    \begin{tabular*}{\textwidth}{@{\extracolsep{\fill}}
        p{0.15\linewidth}p{0.77\linewidth}@{}}
        \toprule
        \textbf{Feature} & \textbf{Operational specification} \\
        \midrule
        Reference frame & Model the coaxial assembly along the global X-axis,
        with the output shaft pointing toward negative X and the terminal box
        above the motor along positive Z. \\
        Motor housing & Create a 100-mm-diameter, 190-mm-long cylindrical barrel
        with stepped front and rear closures, conical transition bands, and two
        shallow circumferential detail bands. \\
        Reduction gearbox & Build a short, bulky octagonal housing approximately
        150 mm across, followed by a lofted transition toward the motor body. \\
        Output flange & Add a circular flange of radius 85 mm and cut six
        uniformly distributed open radial slots through it. \\
        Output shaft & Construct a central boss, stepped shaft base, cylindrical
        shaft blank, reduced tip, and ten longitudinal spline grooves. \\
        Terminal enclosure & Form a rounded rectangular neck, box, rim, and lid;
        add four lid fasteners and three visible cable-entry bosses with recessed
        openings. \\
        Service details & Add separate top and side gearbox service plugs with
        low-profile caps, preserving their attachment to the common housing. \\
        Deliverables & Produce executable \texttt{model.py}, STEP and STL files,
        a \cadstep{} trace, and visual verification against the reference image. \\
        \bottomrule
    \end{tabular*}
    \caption{Refined requirements for the geared electric motor.}
    \label{supp:tab:refined-requirements}
\end{table}

\subsection{Feature Coverage}

The persistent feature tracker initializes eight requested feature groups.
Table~\ref{supp:tab:feature-coverage} consolidates them into seven presentation
rows by combining evidence from the promoted program, successful \cadstep{}
trace, and final visual report. The final program executes twelve \cadstep{}
scopes.

\begin{table}[H]
    \centering
    \small
    \setlength{\tabcolsep}{4pt}
    \renewcommand{\arraystretch}{1.15}
    \begin{tabular*}{\textwidth}{@{\extracolsep{\fill}}
        >{\raggedright\arraybackslash}p{0.16\linewidth}
        >{\raggedright\arraybackslash}p{0.25\linewidth}
        >{\raggedright\arraybackslash}p{0.40\linewidth}
        >{\raggedright\arraybackslash}p{0.12\linewidth}@{}}
        \toprule
        \textbf{Feature} & \textbf{Related CAD step} &
        \textbf{Consolidated evidence} & \textbf{State} \\
        \midrule
        Motor housing & \texttt{motor\_body} & Successful execution; barrel and
        stepped closures visible in multiple views & Verified \\
        Gearbox & \texttt{gearbox\_housing} & Octagonal housing and tapered
        transition execute successfully and are visually present & Verified \\
        Slotted flange & \texttt{output\_flange}, \texttt{flange\_slot\_cut} & Six
        evenly distributed open radial slots are visible & Verified \\
        Stepped shaft & \texttt{output\_shaft}, \texttt{spline\_groove\_cut} & The
        associated steps execute successfully; the raised boss, projecting
        shaft, reduced tip, and ten grooves are visible & Verified \\
        Terminal box & \texttt{terminal\_box}, \texttt{cable\_entry\_bosses},
        \texttt{cable\_opening\_cut} & Rounded enclosure, lid fasteners, and three
        recessed side openings are visible & Verified \\
        Service plugs & \texttt{gearbox\_plugs} & Initially failed; the promoted
        local repair completes without a failed CAD step & Verified \\
        Assembly and export & \texttt{assembly}, \texttt{export\_model} & One final
        assembly is produced; STEP and STL are both verified & Verified \\
        \bottomrule
    \end{tabular*}
    \caption{Feature coverage and evidence for the geared electric motor.}
    \label{supp:tab:feature-coverage}
\end{table}

\subsection{Skill Organization and API Grounding}

The SimpleCADAPI skill organizes documentation hierarchically rather than as a
flat signature list. \texttt{SKILL.md} defines global modeling rules and the
retrieval order, the \texttt{references} directory contains package maps and
reusable workflows, and separate pages document individual APIs and core types.
Listing~\ref{supp:lst:skill-layout} shows the portion relevant to this case.

\noindent\begin{minipage}{\linewidth}
\begin{lstlisting}[style=tracecadpython,language={},numbers=none,
caption={Organization of the SimpleCADAPI documentation skill.},
label={supp:lst:skill-layout}]
skills/simplecadapi/
|-- SKILL.md
|   Global rules, retrieval order, and modeling discipline
`-- references/
    |-- SDK_OVERVIEW.md
    |   Package-level map and public concepts
    |-- MODELING_WORKFLOWS.md
    |   Reusable construction and validation workflows
    `-- docs/
        |-- api/README.md
        |   Index of public API pages
        |-- api/union_rsolid.md
        |   Contract, parameters, failure behavior, and example
        `-- core/solid.md
            Semantics of the Solid type returned by the API
\end{lstlisting}
\end{minipage}

The agent first reads the skill-level rules and API index, then retrieves the
page for the required operation. For \texttt{union\_rsolid}, that page states that
the input may contain individual solids or nested solid sequences, but the
operation must return exactly one \texttt{Solid}. It also documents the
\texttt{clean}, \texttt{glue}, and optional \texttt{tol} parameters, together with
the failure condition for disconnected inputs.

Listing~\ref{supp:lst:api-example} shows how the promoted program applies this
contract. The spatially separated service-plug components remain in a sequence
until final assembly. At that point, each component intersects the common
gearbox body, allowing the complete input to form one connected solid.

\noindent\begin{minipage}{\linewidth}
\begin{lstlisting}[style=tracecadpython,
caption={Concrete use of the \texttt{union\_rsolid} contract in the promoted repair.},
label={supp:lst:api-example}]
gearbox_plugs = [top_plug_stem, top_plug_cap, side_plug_stem, side_plug_cap]

final_model = union_rsolid(
    motor_body, gearbox_body, slotted_output_flange, shaft_assembly,
    finished_terminal_box, gearbox_plugs, tol=0.1, glue=False
)
\end{lstlisting}
\end{minipage}

This organization provides both the global single-solid rule and the local
parameter and failure semantics required for repair, without injecting the
full SDK documentation into the model context.

\subsection{Failure Localization and Localized Repair}

The initial program fails at \texttt{gearbox\_plugs} before export. It directly
fuses the spatially separated top and side plug groups, causing the Boolean
kernel to return two solids and violate the one-solid contract of
\texttt{union\_rsolid}. The execution trace localizes the failed step and
operation without implicating the motor, flange, shaft, or terminal-box code.

Bounded search then evaluates the two candidates in Table~\ref{supp:tab:repair-candidates}. C1 is rejected
before execution because its replacement lacks a valid
\texttt{gearbox\_plugs} step block. C2 makes a 19-line local edit that defers the
disconnected union until final assembly. It executes without a failed
\cadstep{}, exports valid STEP and STL artifacts, and is promoted.

\begin{table}[H]
    \centering
    \small
    \setlength{\tabcolsep}{5pt}
    \begin{tabular*}{\textwidth}{@{\extracolsep{\fill}}lcccc@{}}
        \toprule
        \textbf{Candidate} & \textbf{Patch} & \textbf{Exec.} &
        \textbf{Artifacts} & \textbf{Decision} \\
        \midrule
        C1 & Invalid step replacement & Not run & 0 & Reject \\
        C2 & Deferred Boolean union & Success & 3 & Promote \\
        \bottomrule
    \end{tabular*}
    \caption{Local candidates evaluated for \texttt{gearbox\_plugs}.}
    \label{supp:tab:repair-candidates}
\end{table}

\begin{figure}[H]
    \centering
    \includegraphics[width=\linewidth]{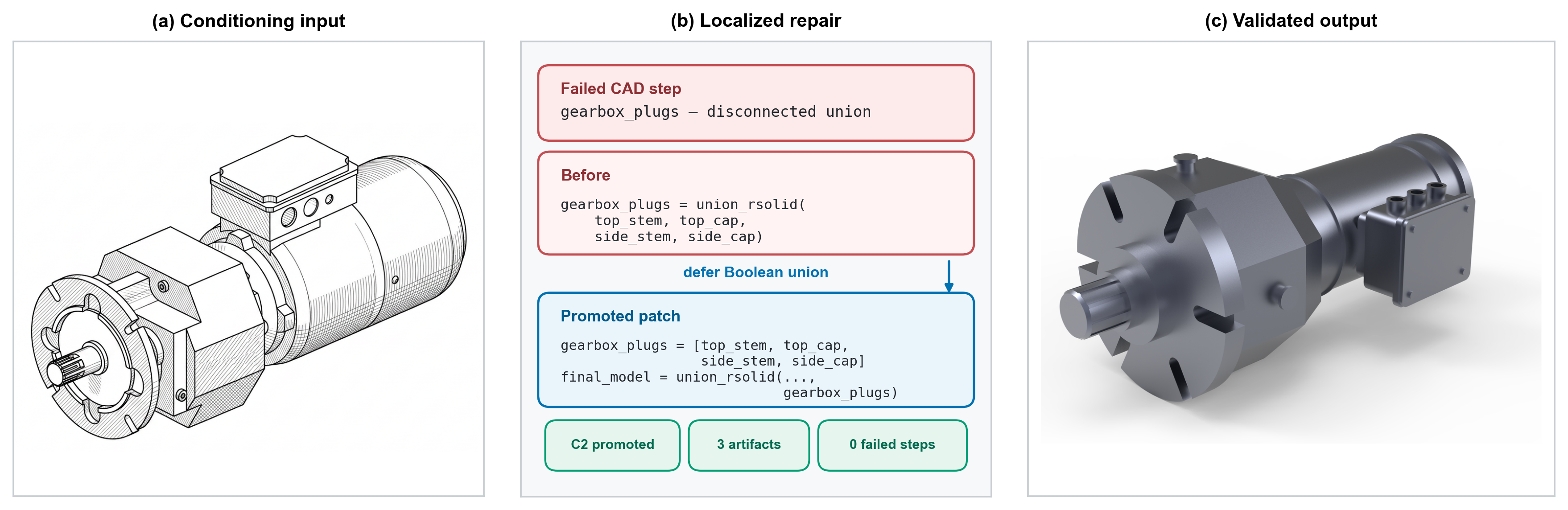}
    \caption{Image-conditioned generation and localized repair of the geared
    electric motor: (a) conditioning sketch, (b) failure localized to
    \texttt{gearbox\_plugs} and the promoted deferred-union patch, and (c) the
    validated output.}
    \label{supp:fig:selected-case}
\end{figure}

Because the failure occurs before the initial program exports a valid STL, no
faithful pre-repair rendering exists. Figure~\ref{supp:fig:selected-case}
therefore places the localized execution evidence and promoted patch between
the input and final output.

After promotion, the final visual verifier accepts the model with 0.94
confidence. It confirms the elongated motor body, faceted gearbox, six-slot
flange, stepped splined shaft, top terminal box, cable-entry openings, and
coaxial alignment, and reports no missing feature or obvious geometric defect.

\subsection{Repair Outcome}

The trajectory records the repair family
\texttt{deferred\_\allowbreak boolean\_\allowbreak union}: disconnected detail groups
remain separate until they can be merged with a common intersecting parent
during assembly.

\subsection{Example of an Induced Repair Skill}

A separate foot-mounted electric-motor run illustrates skill induction. The
run completed without an execution error and produced an STL file,
but downstream rendering found no valid triangles or plottable vertices. After
the export path was repaired and the result passed execution and visual checks,
\method{} induced the skill
\texttt{repair\_\allowbreak empty\_\allowbreak stl\_\allowbreak from\_\allowbreak
cad\_\allowbreak assembly\_\allowbreak export}.
Table~\ref{supp:tab:induced-skill} summarizes the persisted record.

\begin{table}[H]
    \centering
    \small
    \setlength{\tabcolsep}{4pt}
    \renewcommand{\arraystretch}{1.15}
    \begin{tabular*}{\textwidth}{@{\extracolsep{\fill}}
        >{\raggedright\arraybackslash}p{0.15\linewidth}
        >{\raggedright\arraybackslash}p{0.77\linewidth}@{}}
        \toprule
        \textbf{Field} & \textbf{Persisted content} \\
        \midrule
        Failure signature & Nominally successful \texttt{export\_model} step,
        but the STL contains no renderable vertices or triangles. \\
        Diagnosis & Artifact existence and a zero process return code do not
        establish successful mesh export; an assembly or container may not have
        been triangulated into usable geometry. \\
        Repair policy & Collect valid visible solids, triangulate them
        explicitly, combine only nonempty meshes, and reopen the STL to verify
        its size, vertex and triangle counts, and finite bounds. \\
        Success evidence & The BRep exports remain valid, while the repaired
        STL is nonempty and renders as a coherent motor in six views. \\
        Reusable context & Multi-solid or assembly-like CAD models whose BRep
        export succeeds but whose assembly-level STL is empty. \\
        \bottomrule
    \end{tabular*}
    \caption{Persisted fields of an induced export-repair skill.}
    \label{supp:tab:induced-skill}
\end{table}

The record separates applicability, diagnosis, repair policy, and validation
evidence.

\subsection{Final Validation}

The promoted geared-motor program contains twelve \cadstep{} scopes that map
to the coverage entries in Table~\ref{supp:tab:feature-coverage}. It exports the
final STEP and STL models and passes the final visual verification.

\subsection{Mapping to FreeCAD}

\begin{figure}[H]
    \centering
    \includegraphics[width=0.95\linewidth]{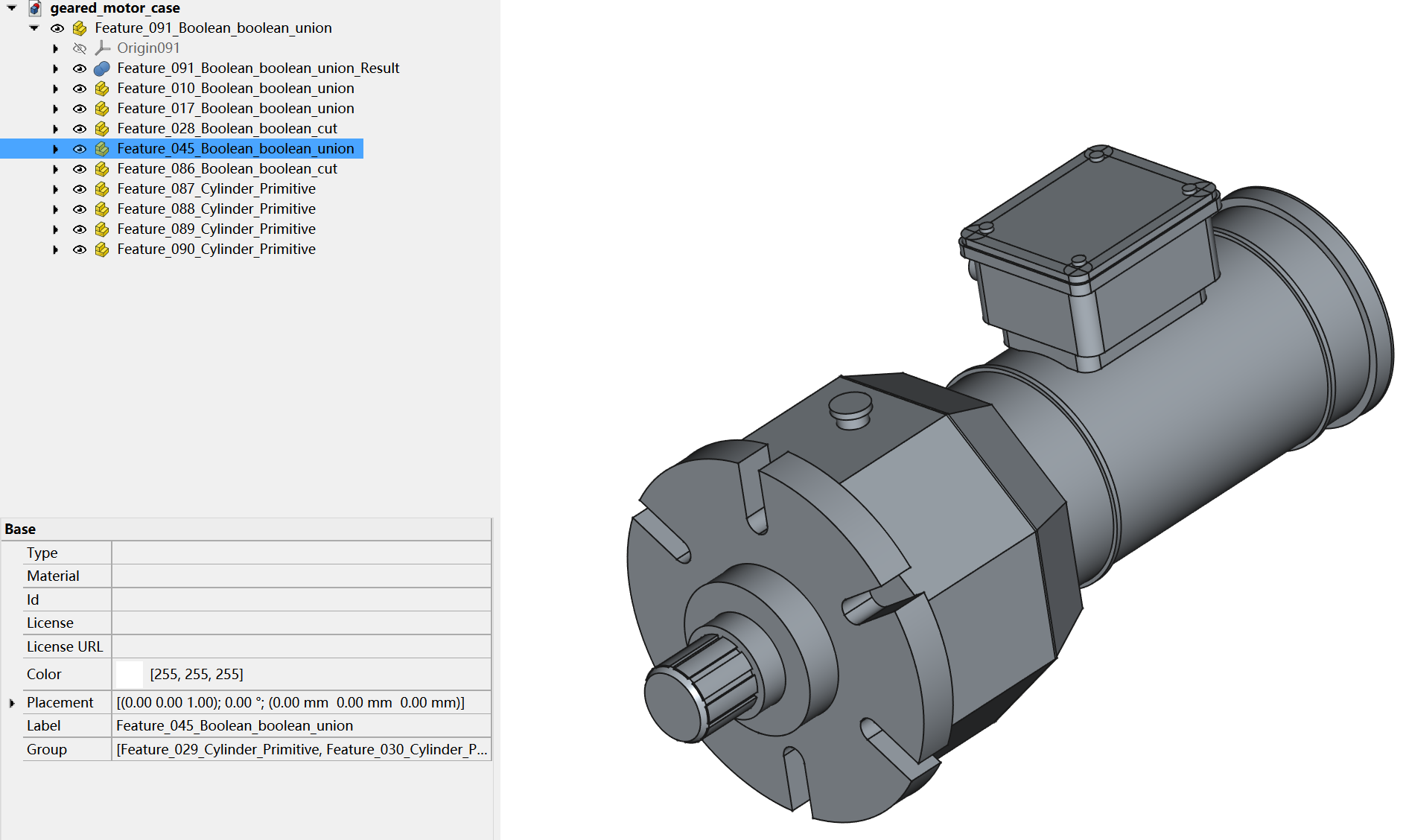}
    \caption{The validated geared-motor model reconstructed in FreeCAD, with
    its mapped feature tree shown on the left.}
    \label{supp:fig:freecad-conversion}
\end{figure}

A conversion script maps the validated sequence of SimpleCADAPI operations to a
FreeCAD-compatible Python macro. Executing the macro reconstructs the model in
FreeCAD for parameter inspection, downstream editing, and export.
Figure~\ref{supp:fig:freecad-conversion} shows the resulting
editable FreeCAD model and feature hierarchy.

\end{document}